\documentclass[]{fairmeta}
\usepackage{xcolor}
\usepackage[utf8]{inputenc}
\usepackage{multirow}
\usepackage{nameref}

\usepackage[utf8]{inputenc} 
\usepackage[T1]{fontenc}    
\usepackage{url}            
\usepackage{amsfonts}       
\usepackage{nicefrac}       
\usepackage{microtype}      
\usepackage{xcolor}         
\usepackage{wrapfig}        
\usepackage[nohyperlinks]{acronym} 
\usepackage{bbm}            
\usepackage{xspace}         
\usepackage{enumitem}       
\usepackage{bm}             
\usepackage{multirow}

\usepackage[textsize=tiny]{todonotes}

\usepackage{color}

\usepackage{makecell} 
\usepackage{fontawesome5}  
\usepackage{amssymb}
\usepackage{adjustbox}
\usepackage{amsthm}
\usepackage{amsmath}        
\usepackage{listings}       
\usepackage{mathtools}      

\usepackage{booktabs}       
\usepackage{nicematrix}
\usepackage{bm}             

\newcommand{\our}{Q-NeuS}

\makeatletter
\DeclareRobustCommand\onedot{\futurelet\@let@token\@onedot}
\def\@onedot{\ifx\@let@token.\else.\null\fi\xspace}

\makeatother

\definecolor{safe}{HTML}{69A831}   
\definecolor{unsafe}{HTML}{A657D6} 
\definecolor{methodfg}{HTML}{FF8800}
\definecolor{methodbg}{RGB}{255,244,227}
\definecolor{lightgray}{HTML}{f0f0f0}
\definecolor{mypurple}{HTML}{A13FDB} 
\definecolor{mypink}{HTML}{E14B9E} 
\definecolor{myblue}{HTML}{456AFF} 
\definecolor{myred}{HTML}{F20D0D} 
\definecolor{mygreen}{HTML}{4ED65C} 

\acrodef{lrm}[LRM]{large reasoning model}
\acrodef{cot}[CoT]{chain-of-thought}
\acrodef{sft}[SFT]{supervised finetuning}
\acrodef{rlhf}[RLHF]{reinforcement learning from human feedback}
\acrodef{dapo}[DAPO]{decouple clip and dynamic sampling policy optimization}
\acrodef{grpo}[GRPO]{group relative policy optimization }
\acrodef{rl}[RL]{reinforcement learning}
\acrodef{rlvr}[RLVR]{Reinforcement Learning with Verifiable Rewards}
\acrodef{ipr}[IPR]{iterative prefill reset}
\acrodef{mdp}[MDP]{markov decision process}
\acrodef{llm}[LLM]{large language model}

\begin{document}
\title{Query-Adaptive Neuron Steering for LLMs}

\author[1]{Yanxi Pan}
\author[2]{Jingwen Wu}
\author[1]{Yan Ma}
\author[3]{Xiang Chen}
\author[4]{Xinyi Zeng}
\author[5]{Xuetong Wang}
\author[6]{Yunyan Zhang}
\author[7]{Yue Zhang}
\author[5, *, \ddagger]{Yu Tian}
\author[5,*]{Jianfeng Pan}
\author[4]{Hongyi Zhou}

\affiliation[1]{Northeastern University}
\affiliation[2]{Washington University in St. Louis}
\affiliation[3]{Nanjing University of Aeronautics and Astronautics}
\affiliation[4]{Tsinghua University}
\affiliation[5]{360 Security Capability Center}
\affiliation[6]{Tencent}
\affiliation[7]{Beijing University of Posts and Telecommunications}
\contribution[*]{Corresponding author}
\contribution[\ddagger]{Project Leader}

\abstract{Effectively eliciting the capabilities of large language models (LLMs) at inference time remains an important challenge.
Existing studies address this challenge through context engineering, which shapes user queries, or fixed steering, which directly intervenes in model computation. However, predefined contexts and fixed intervention strategies may not adequately accommodate the diverse knowledge and reasoning requirements of individual queries. This raises a natural question: \textbf{Can internal
steering in LLMs be made query-adaptive?} 
In this work, we investigate whether query-induced internal activations can guide adaptive steering in LLMs. Our pilot studies reveal subject- and layer-dependent neuron activation patterns and show that amplifying subject-relevant neurons can improve model performance.
Based on these findings, we propose \textbf{\our{}}, a query-adaptive neuron steering framework that dynamically determines which neurons to modulate and by how much for each input. Specifically, \our{} constructs a compact subject-guided candidate neuron space and employs a lightweight Query-Adaptive Gate to generate neuron-wise modulation coefficients. 
To learn effective intervention strategies, we introduce \textbf{Neu-GRPO}, which adapts group-relative optimization to train the Query-Adaptive Gate using the relative performance of multiple neuron intervention configurations. The base LLM remains frozen throughout training and inference, and no task or subject labels are required at inference time.
Extensive experiments across multiple LLMs and benchmarks demonstrate that \our{} consistently improves model performance and exhibits broad applicability across model scales and strong cross-benchmark transferability.
}

\date{\today}
\correspondence{Yu Tian \email{tianyu181@mails.ucas.ac.cn}, Yanxi Pan \email{panyx1@mails.neu.edu.cn}}

\metadata[Code]{\url{https://github.com/YanxiP/Q-NeuS}}

\maketitle
\section{Introduction}
\label{section: intro}

\begin{figure*}[!t]
    \centering
    \includegraphics[width=\textwidth]{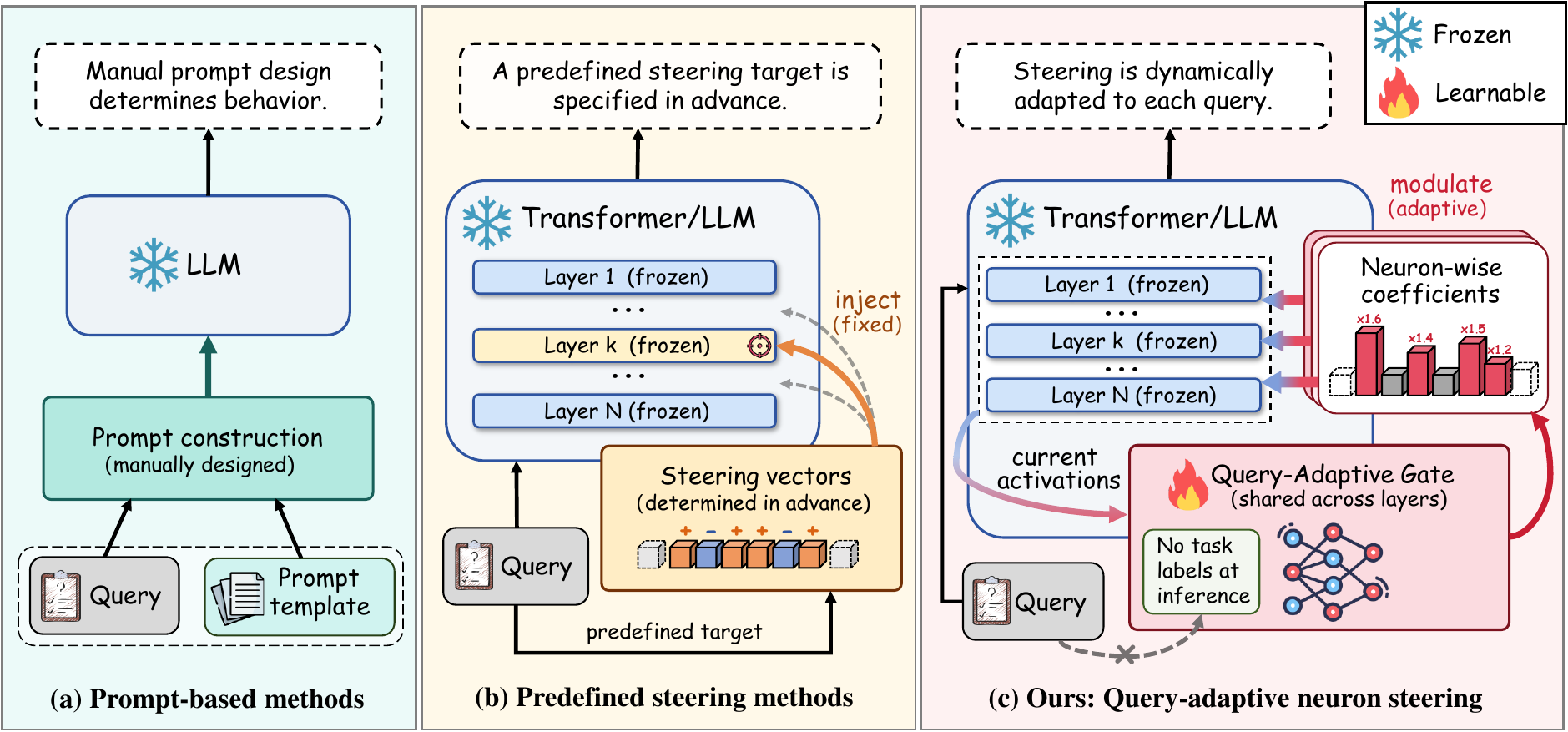}
    \caption{Comparison of paradigms for eliciting latent capabilities in LLMs. Prompt-based methods modify model inputs, predefined steering applies fixed internal interventions, while \our{} dynamically adapts neuron-level interventions to each query.}
    \label{fig:introduction}
\end{figure*}

Exploring and eliciting the latent capabilities of Large Language Models (LLMs) is crucial for understanding and improving their latent potential.
\citep{greenblatt2024stress,donoway2025quantifying,hofstatter2025the}.
Existing studies largely follow two directions (As illustrated in Fig. \ref{fig:introduction}). The first operates on model inputs through context engineering 
\citep{wang2025demonstration,ramnath2025systematic,zhao2026gradient}.
While effective, these methods often rely on manually designed or task-specific contexts and are sensitive to how well the provided context matches each query. The second directly manipulates internal representations through steering
\citep{rimsky2024steering,stolfo2025improving,wang2025beyond}. 
However, existing steering methods typically target predefined tasks, capabilities, or behaviors, with the steering target or direction identified in advance. Such predefined interventions are difficult to generalize to heterogeneous queries that may require different knowledge and reasoning processes. Despite operating at different levels, both paradigms still rely on an intervention strategy specified in advance rather than adapting it to the current query. This raises a natural question: \textbf{Can internal steering in LLMs be made query-adaptive?} 

To investigate this question, we first examine whether the internal activations of LLMs provide informative signals for query-adaptive intervention. We conduct a series of preliminary experiments to characterize neuron activation patterns across various domains and evaluate the effects of intervening on highly activated neurons. We observe that 1) highly activated neurons exhibit clear subject- and layer-dependent structures, and 2) amplifying subject-relevant neurons can improve model performance, while the intervention effects vary substantially across layers and subjects. These findings indicate that neuron activations encode task-relevant information that can guide internal intervention, while revealing the limitations of fixed steering strategies for heterogeneous queries. 

Based on these pilot results, we introduce \our{}, a query-adaptive neuron steering framework that identifies and modulates neurons according to the internal state induced by each query. Specifically, we first construct a subject-guided candidate neuron space from training data to constrain the intervention search space. We then introduce a lightweight Query-Adaptive Gate that takes the  internal activations as input and dynamically generates neuron-wise modulation coefficients, allowing different queries to induce different steering patterns. To learn effective intervention strategies, we further propose Neu-GRPO, which adapts GRPO to the neuron intervention setting and optimizes the Query-Adaptive Gate by comparing the relative performance of multiple neuron intervention configurations. During inference, the method requires no task or domain labels, and the modulation pattern is determined solely by the internal activation state of the current query.
Extensive experiments on seven LLMs ranging from 3B to 32B parameters across MMLU-Pro, SuperGPQA, and WildSci show that \our{} improves performance across all evaluated models and benchmarks. Notably, \our{} trained only on MMLU-Pro transfers effectively to multiple unseen benchmarks, demonstrating broad applicability across model scales and strong cross-benchmark transferability.
\section{Related Work}
\label{sec:related_work}

\noindent\textbf{Activation Steering for LLMs.} Activation steering has emerged as an effective approach for controlling LLM behavior by directly modifying internal activations without updating the base model parameters 
\citep{li2023inference,rimsky2024steering,stolfo2025improving}. 
Prior work has demonstrated its effectiveness in behavior control and reasoning enhancement
\citep{hojer2025improving}, 
typically using techniques such as Contrastive Activation Addition (CAA) \citep{rimsky2024steering} or Sparse Autoencoders (SAEs) 
\citep{cho2025toward} 
to construct or identify steerable representations. More recent studies explore conditional and adaptive 
steering \citep{lee2025programming,ye2026riser,zheng2026dpn}.
However, these methods remain constrained by a preconstructed intervention space, where the available steering directions or primitives are identified in advance. Unlike these methods, \our{} performs query-adaptive neuron modulation based on query-induced activations, enabling fine-grained intervention beyond predefined steering spaces.

\noindent\textbf{RL-based Policy Optimization for LLMs.} Reinforcement learning (RL) has been widely used to optimize LLM generation policies 
\citep{ouyang2022training,shao2024deepseekmath,guo2025deepseek}. 
PPO-style RLHF typically relies on a learned value function for advantage
estimation, introducing additional memory and computational overhead
\citep{schulman2017proximal,ouyang2022training}. 
GRPO removes the critic by estimating advantages from the relative rewards of multiple outputs for the same prompt, thereby reducing training overhead 
\citep{shao2024deepseekmath}.
Recent studies have extended GRPO to adaptive routing and other settings 
\citep{ye2026riser}. 
Different from methods that optimize the LLM policy itself, we keep the base LLM frozen and adapt GRPO to optimize the Query-Adaptive Gate according to the relative performance of different neuron intervention configurations.
\section{Preliminary Study}
\label{section: architecture}

\begin{figure*}[!t]
    \centering

    \begin{minipage}[t]{0.65\textwidth}
        \vspace{0pt}
        \centering
        \includegraphics[
            width=\linewidth,
            keepaspectratio
        ]{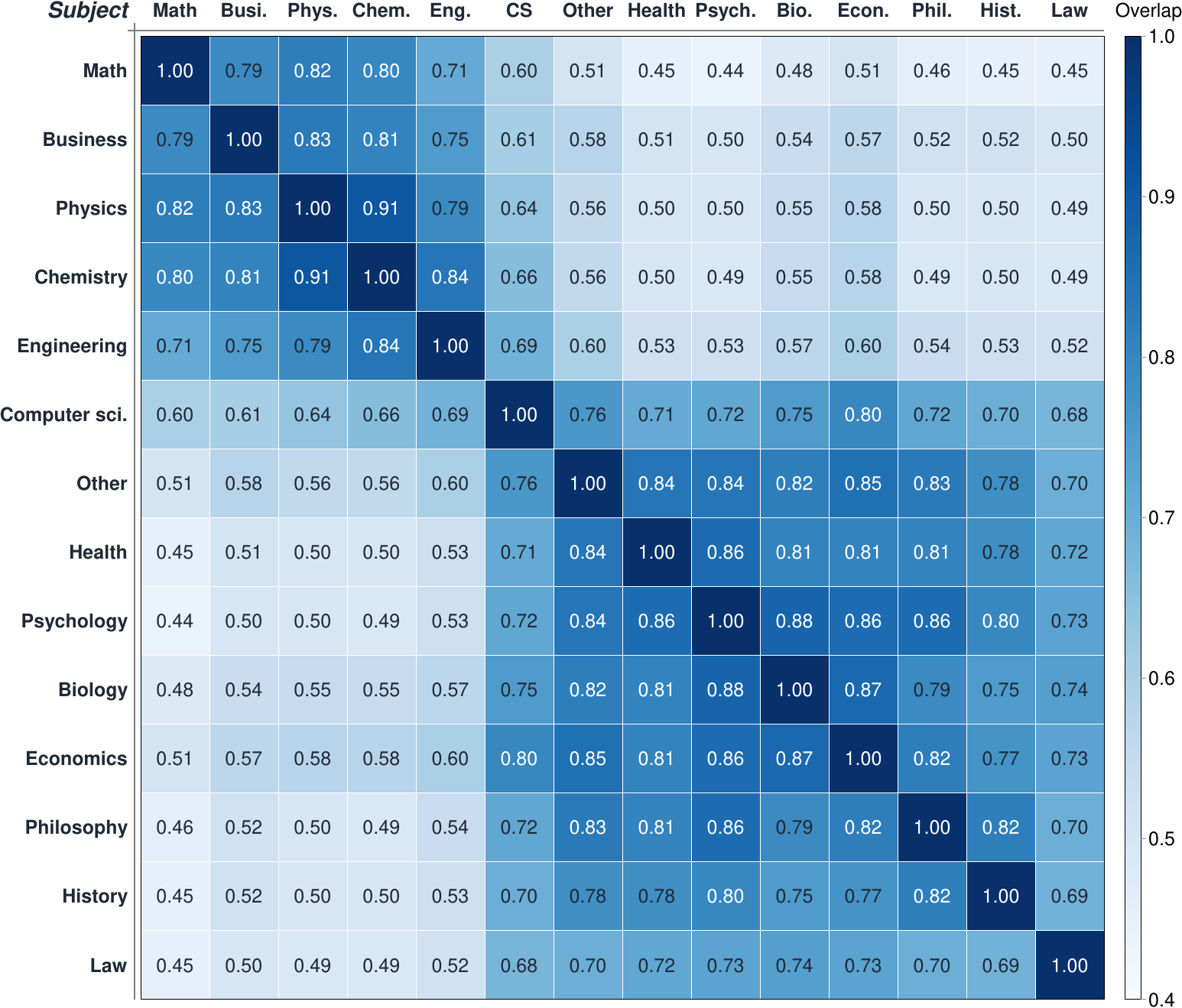}
        \captionof{figure}{
            Overlap matrix of different subjects'
            Top-500 activation neurons in layer 25.
        }
        \label{fig:subject-overlap}
    \end{minipage}
    \hfill
    \begin{minipage}[t]{0.3\textwidth}
        \vspace{0pt}
        \centering
        \captionof{table}{
            Subject-wise performance on MMLU-Pro with
            Qwen2.5-3B-Instruct.
        }
        \label{tab:mmlu_domain_pre}
        \small
        \setlength{\tabcolsep}{1.5pt}
        \renewcommand{\arraystretch}{1.16}
        \begin{tabular}{lccc}
            \toprule
            \textbf{Subject}
            & \textbf{Baseline}
            & \textbf{Fix}
            & $\boldsymbol{\Delta}$ \\
            \midrule
            Math      & 20.66 & 21.40
                & \textcolor{green!50!black}{+0.74} \\
            Busi.         & 20.75 & 20.13
                & \textcolor{red!70!black}{-0.63} \\
            Phys.          & 19.54 & 22.61
                & \textcolor{green!50!black}{+3.07} \\
            Chem.        & 12.78 & 18.06
                & \textcolor{green!50!black}{+5.29} \\
            Eng.      & 28.21 & 33.33
                & \textcolor{green!50!black}{+5.13} \\
            CS & 30.49 & 34.15
                & \textcolor{green!50!black}{+3.66} \\
            Other            & 36.02 & 35.48
                & \textcolor{red!70!black}{-0.54} \\
            Health           & 44.85 & 46.06
                & \textcolor{green!50!black}{+1.21} \\
            Psych.       & 49.69 & 52.17
                & \textcolor{green!50!black}{+2.48} \\
            Bio.       & 59.31 & 57.93
                & \textcolor{red!70!black}{-1.38} \\
            Econ.        & 51.76 & 51.18
                & \textcolor{red!70!black}{-0.59} \\
            Phil.       & 33.66 & 28.71
                & \textcolor{red!70!black}{-4.95} \\
            Hist.          & 46.75 & 46.75
                & 0.00 \\
            Law              & 22.62 & 23.53
                & \textcolor{green!50!black}{+0.90} \\
            \midrule
            \textbf{Overall}
                & \textbf{31.56}
                & \textbf{32.92}
                & \textbf{\textcolor{green!50!black}{+1.36}} \\
            \bottomrule
        \end{tabular}
    \end{minipage}
\end{figure*}

In this section, we investigate whether LLM activations support query-adaptive intervention by relating subject-level queries to neuron activations. We conduct experiments with Qwen2.5-3B-Instruct on all 14 subjects of MMLU-Pro, splitting each subject into training/validation/test sets (70\%/10\%/20\%) for neuron identification, layer selection, and final evaluation, respectively. For each layer, we extract the MLP activations at the final prompt token immediately before the output projection and rank neurons by their mean activation over the training examples. We then multiply the activations of the top 100 neurons by 1.5 and those of the next 400 neurons by 1.2, while leaving the remaining neurons unchanged. We independently intervene on each layer and retain those that improve validation accuracy. Finally, all selected layers are jointly intervened on and evaluated on the held-out test set. More details are provided in Appendix \ref{app:preliminary}.

\noindent\textbf{Observation 1. Different subjects exhibit distinct neuron activation patterns.} We examine the organization of highly activated neurons across subjects by selecting the top 500 neurons in each layer and computing pairwise overlap. As shown in Fig. \ref{fig:subject-overlap}, semantically related subjects exhibit higher overlap and form clear clusters, while substantial sharing remains even across dissimilar subjects. This suggests that LLMs rely on both shared computational components and subject-specific neuron subsets. As shown in Appendix \ref{app:preliminary}, this organization varies across depth: shallow layers are largely shared across subjects, intermediate layers exhibit stronger subject-specific structure, and the distinction weakens again in deeper layers. These results indicate that neuron organization is both subject-dependent and layer-dependent, motivating layer-aware and query-adaptive intervention.

\noindent\textbf{Observation 2. Amplifying subject-relevant neurons improves model performance.} As shown in Tab. \ref{tab:mmlu_domain_pre}, jointly intervening on the 21 positive-gain layers selected on the validation set improves test accuracy from 31.56\% to 32.92\% (+1.36 pp), demonstrating that selectively amplifying subject-relevant neurons can effectively improve model performance. However, the gains vary substantially across layers and subjects, with some subjects even experiencing performance degradation under joint intervention. These results suggest that while neuron-level intervention is effective, a fixed intervention strategy cannot consistently accommodate the heterogeneous requirements of different queries, motivating query-adaptive neuron steering.

\noindent\textbf{Motivation.} These results show that modulating subject-relevant neurons can effectively improve model performance, while the optimal intervention varies across layers and subjects. However, our pilot study relies on manually specified neuron counts, fixed modulation coefficients, and predetermined layer selection, limiting its ability to adapt to individual queries. To address this limitation, we introduce \our{}, which employs a learnable gating mechanism to dynamically determine which neurons to modulate and by how much according to the internal activation state of each query.
\section{\our{}}
\label{sec:method}

\begin{figure*}[!t]
    \centering
    \includegraphics[width=\textwidth]{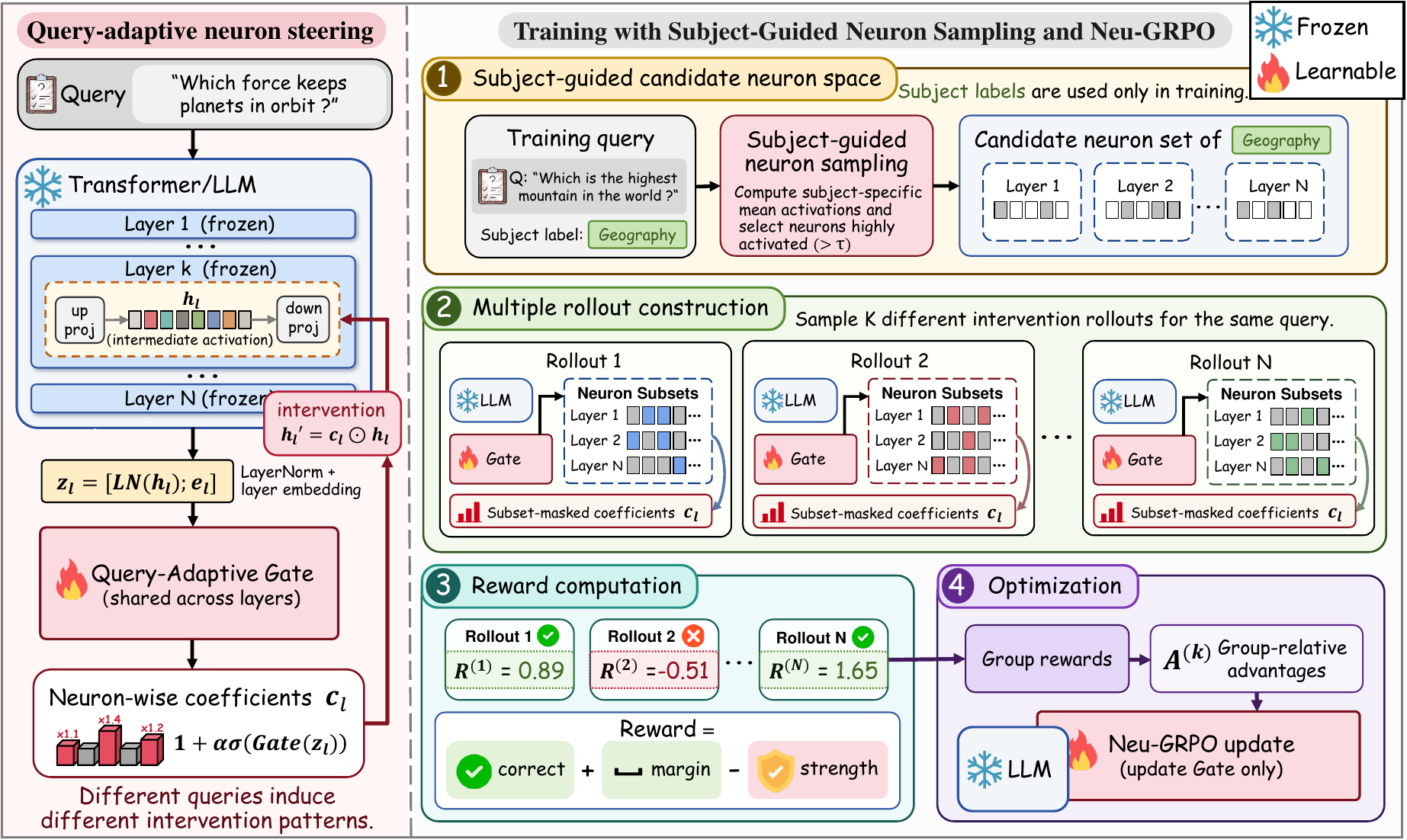}
    \caption{
    Overview of \our{}. \our{} generates query-adaptive neuron-wise modulation coefficients from query-induced activations. During training, subject-guided neurons form intervention rollouts evaluated by correctness, margin, and intervention-strength rewards. Neu-GRPO optimizes the Query-Adaptive Gate with group-relative feedback while keeping the base LLM frozen.}
    \label{method}
\end{figure*}

Based on these pilot findings, we introduce \our{}, a query-adaptive neuron steering framework. During training, subject-guided candidate neurons are explored through multiple rollouts, and a Query-Adaptive Gate is optimized with Neu-GRPO using their relative performance. At inference time, the Query-Adaptive Gate dynamically modulates all FFN neurons across different Transformer layers based on the internal activations induced by each query, without requiring subject labels.

\subsection{Query-Adaptive Neuron Gating}
\label{sec:query_adaptive_gating}

\noindent\textbf{Neuron Specification.}
Most LLMs employ a feed-forward network (FFN) in each Transformer layer.
Given the hidden state $\mathbf{x}_l$ of the last input token at layer $l$, the FFN can be written as:
\begin{equation}
\mathbf{h}_l =
\operatorname{SiLU}
\left(
W_l^{\mathrm{gate}}\mathbf{x}_l
\right)
\odot
\left(
W_l^{\mathrm{up}}\mathbf{x}_l
\right),
\qquad
\mathbf{y}_l =
W_l^{\mathrm{down}}\mathbf{h}_l ,
\end{equation}
where $W_l^{\mathrm{gate}}$, $W_l^{\mathrm{up}}$, and $W_l^{\mathrm{down}}$ denote the gate, up, and output projections, respectively. The vector $\mathbf{h}_l \in \mathbb{R}^{d_{\mathrm{ff}}}$ denotes the FFN intermediate activation before the output projection, where each dimension is treated as an FFN neuron. \our{} directly modulates $\mathbf{h}_l$ before the output projection.

\noindent\textbf{Query-Adaptive Neuron Gating.}
Motivated by our preliminary findings, we introduce a lightweight Query-Adaptive Gate that generates neuron-wise modulation coefficients from query-induced internal activations. To capture layer-specific information, the shared Query-Adaptive Gate takes the normalized activation $\tilde{\mathbf{h}}_l$ together with a learnable layer embedding $\mathbf{e}_l$:
\begin{equation}
\tilde{\mathbf{h}}_l
=
\operatorname{LayerNorm}(\mathbf{h}_l),
\qquad
\mathbf{z}_l
=
[\tilde{\mathbf{h}}_l;\mathbf{e}_l].
\end{equation}
The Query-Adaptive Gate produces gating logits through a two-layer feedforward network:
\begin{equation}
\mathbf{g}_l
=
W_2
\operatorname{SiLU}
\left(
W_1\mathbf{z}_l+\mathbf{b}_1
\right)
+\mathbf{b}_2,
\end{equation}
where $W_1$, $W_2$ and
$\mathbf{b}_1$ and $\mathbf{b}_2$ are learnable weights and biases, respectively. The resulting logits are mapped to neuron-wise modulation coefficients:
\begin{equation}
\mathbf{c}_l =
\mathbf{1}
+
\alpha \sigma(\mathbf{g}_l),
\end{equation}
where $\sigma(\cdot)$ denotes the sigmoid function and $\alpha$ controls the maximum modulation strength.
The intermediate activations are then modulated as:
\begin{equation}
\hat{\mathbf{h}}_l
=
\mathbf{c}_l
\odot
\mathbf{h}_l,
\end{equation}
the modulated activation $\hat{\mathbf{h}}_l$ is passed to the original output projection.

The Query-Adaptive Gate is shared across Transformer layers, while the layer embedding $\mathbf{e}_l$ enables layer-specific modulation. Conditioned on query-induced internal activations, it generates neuron-wise coefficients that adapt both the intervention pattern and modulation strength to each query, without requiring task or subject labels. 
It provides fine-grained and parameter-efficient adaptation to heterogeneous queries, with detailed parameter analysis provided in Appendix~\ref{app:parameter_efficiency}.

\subsection{Neu-GRPO}
\label{sec:gate_training}

To learn query-adaptive neuron modulation policies of \our{}, we introduce \textbf{Neu-GRPO}, which adapts GRPO to neuron-level intervention. For each training query, we construct multiple neuron intervention configurations and evaluate their relative task performance. The resulting group-relative feedback is then used to optimize \our{}, while keeping the base LLM frozen.

\noindent\textbf{Subject-Guided Neuron Sampling.}
Our preliminary study shows that highly activated neurons exhibit subject-specific patterns and provide effective targets for intervention. We leverage this property to construct a compact candidate neuron space, reducing the search space for subsequent query-adaptive modulation. Given $N_c$ training examples from subject $c$, the mean activation of neuron $j$ in layer $l$ is computed as:
\begin{equation}
\bar{h}_{c,l,j}
=
\frac{1}{N_c}
\sum_{i=1}^{N_c}
h_{l,j}^{(i)}.
\end{equation}

We define the subject- and layer-specific candidate set $S_{c,l}$ as:
\begin{equation}
S_{c,l}
=
\left\{
j \mid \bar{h}_{c,l,j} > \tau
\right\},
\end{equation}
where $\tau$ is an activation threshold. Unlike the fixed-size neuron selection used in our preliminary study, this formulation allows the candidate set to adapt to different subjects and layers.

For each training query from subject $c$, we construct $K$ intervention rollouts. In rollout $k$, a subset of candidate neurons $M_l^{(k)}$ is sampled at each layer:
\begin{equation}
M_l^{(k)} \subseteq S_{c,l},
\qquad
|M_l^{(k)}|
=
\left\lfloor \rho |S_{c,l}| \right\rfloor,
\qquad
k=1,\ldots,K,
\end{equation}
where $\rho$ denotes the neuron sampling ratio.
We sample diverse neuron subsets across rollouts, while \our{} adaptively determines their modulation strengths from query-induced activations.

\noindent\textbf{Group-Relative Gate Optimization.} For each rollout $k$, \our{} multiplicatively generates neuron-wise modulation coefficients and applies them to the sampled neurons $M_l^{(k)}$, resulting in a query-specific intervention configuration. Let $\pi_{\phi}^{(k)}(\cdot \mid q)$ denote the output distribution of the modulated model for query $q$, where $\phi$ denotes the parameters of \our{}.
We evaluate each rollout using a reward that considers task correctness, prediction confidence, and intervention strength:
\begin{equation}
R^{(k)}
=
R_{\mathrm{correct}}^{(k)}
+
\lambda_m R_{\mathrm{margin}}^{(k)}
-
\lambda_s R_{\mathrm{strength}}^{(k)},
\end{equation}
where $\lambda_m$ and $\lambda_s$ control the contributions of the margin reward and intervention-strength penalty, respectively. The correctness reward indicates whether the predicted answer is correct.
To capture the model's preference for the correct answer, we further define:
\begin{equation}
R_{\mathrm{margin}}^{(k)}
=
\log \pi_{\phi}^{(k)}(y \mid q)
-
\max_{a' \neq y}
\log \pi_{\phi}^{(k)}(a' \mid q),
\end{equation}
where $y$ is the ground-truth answer. We additionally penalize
unnecessarily strong interventions:
\begin{equation}
R_{\mathrm{strength}}^{(k)}
=
\frac{1}{|\mathcal{V}^{(k)}|}
\sum_{l \in \mathcal{V}^{(k)}}
\frac{1}{|M_l^{(k)}|}
\sum_{j \in M_l^{(k)}}
\left(c_{l,j}^{(k)}-1\right)^2,
\end{equation}
where $\mathcal{V}^{(k)}$ is the set of layers containing sampled
neurons in rollout $k$. We optimize the Query-Adaptive Gate using GRPO and the relative rewards of the $K$ intervention rollouts.

During training, the base LLM remains frozen and only the Query-Adaptive Gate is optimized, while subject labels are used solely to construct candidate neuron sets. At inference time, the learned Query-Adaptive Gate generates neuron-wise modulation coefficients from query-induced internal activations without requiring task or subject labels. This enables subject-guided policy learning during training while retaining fully query-adaptive, label-free intervention at inference time.
\section{Experiment}

\label{sec:experiments}

\subsection{Experimental Setup}
\label{sec:experimental_setup}

\noindent\textbf{Models.}
We evaluate \our{} on seven LLMs spanning different model sizes and families: Qwen2.5-3B \citep{qwen2025qwen25technicalreport}, Qwen3-4B, Qwen3-8B, Qwen3-14B, Qwen3-32B \citep{yang2025qwen3}, Ministral3-8B, and Ministral3-14B \citep{liu2026ministral}.

\noindent\textbf{Datasets and Evaluation.}
We use MMLU-Pro
\citep{wang2024mmlu}
, SuperGPQA
\citep{du2026supergpqa}
, and WildSci
\citep{liu2026wildsci}
as our primary benchmarks. For in-domain evaluation, we partition each benchmark into training (70\%),
validation (10\%), and test (20\%) splits, train Q-NeuS on the training split, and report results on the held-out test split. To assess cross-benchmark generalization, we directly evaluate the model trained on MMLU-Pro on GPQA-Main, GPQA-Diamond
\citep{rein2023gpqa}
, SuperGPQA, LiveBench
\citep{white2025livebench}
, AGIEval
\citep{zhong2024agieval}
, GSM8K
\citep{cobbe2021training}
, and ARC-Challenge
\citep{clark2018think}
, without using any training data from these target benchmarks. Accuracy is used as the primary evaluation metric.

\noindent\textbf{Baselines.} 
We compare Q-NeuS with three baselines. \textbf{Base} is the original LLM without intervention. \textbf{Fix} follows our preliminary study, selecting highly activated neurons from training-set mean activations, applying predefined modulation coefficients, and selecting intervention layers on the validation set. \textbf{GRPO} optimizes LoRA parameters \citep{hu2021lora} with GRPO \citep{shao2024deepseekmath}. All methods use the same base model and evaluation protocol whenever applicable.

\noindent\textbf{Implementation Details.}
We set the candidate neuron threshold to $\tau=0.1$. The Query-Adaptive Gate uses a hidden dimension of $2{,}048$ and a layer embedding dimension of $32$, with a maximum modulation coefficient of $1.7$. For each query, we construct $K=4$ rollouts and sample $60\%$ of the candidate neurons in each layer. In Neu-GRPO, we set $\lambda_m=1.0$ and $\lambda_s=0.1$ for the margin reward and intervention strength penalty, respectively. \our{} is trained with Adam at a learning rate of $1\times10^{-4}$ for up to $3$ epochs. All main experiments are conducted on NVIDIA A800 GPUs.

\subsection{Main Results}
\label{sec:main_results}

\begin{table*}[!t]
    \centering
    \caption{
        Performance  of Q-NeuS across different LLMs on MMLU-Pro, SuperGPQA, and WildSci.
    }
    \label{tab:main_results}
    \small
    \setlength{\tabcolsep}{4.2pt}
    \renewcommand{\arraystretch}{1.16}

    \begin{tabular}{lccccccccc}
        \toprule

        \multirow{2}{*}{\textbf{Model}}
        & \multicolumn{3}{c}{\textbf{MMLU-Pro}}
        & \multicolumn{3}{c}{\textbf{SuperGPQA}}
        & \multicolumn{3}{c}{\textbf{WildSci}} \\

        \cmidrule(lr){2-4}
        \cmidrule(lr){5-7}
        \cmidrule(lr){8-10}

        &
        \textbf{Base} & \textbf{Q-NeuS} & $\boldsymbol{\Delta}$
        &
        \textbf{Base} & \textbf{Q-NeuS} & $\boldsymbol{\Delta}$
        &
        \textbf{Base} & \textbf{Q-NeuS} & $\boldsymbol{\Delta}$
        \\

        \midrule

        Qwen2.5-3B
        & 31.43 & 38.37 & \textcolor{green!50!black}{+6.94}
        & 19.58 & 24.37 & \textcolor{green!50!black}{+4.79}
        & 33.95 & 39.51 & \textcolor{green!50!black}{+5.56} \\

        Qwen3-4B
        & 44.86 & 45.93 & \textcolor{green!50!black}{+1.07}
        & 26.29 & 29.79 & \textcolor{green!50!black}{+3.50}
        & 43.20 & 45.52 & \textcolor{green!50!black}{+2.32} \\

        Qwen3-8B
        & 44.20 & 48.53 & \textcolor{green!50!black}{+4.33}
        & 26.83 & 30.95 & \textcolor{green!50!black}{+4.12}
        & 42.91 & 45.78 & \textcolor{green!50!black}{+2.87} \\

        Qwen3-14B
        & 50.89 & 54.03 & \textcolor{green!50!black}{+3.14}
        & 30.35 & 35.46 & \textcolor{green!50!black}{+5.11}
        & 46.91 & 49.58 & \textcolor{green!50!black}{+2.67} \\

        Qwen3-32B
        & 54.56 & 59.11 & \textcolor{green!50!black}{+4.55}
        & 33.97 & 38.73 & \textcolor{green!50!black}{+4.76}
        & 49.64 & 52.56 & \textcolor{green!50!black}{+2.92} \\

        \midrule

        Ministral3-8B
        & 44.82 & 49.03 & \textcolor{green!50!black}{+4.21}
        & 24.50 & 28.68 & \textcolor{green!50!black}{+4.18}
        & 45.05 & 50.01 & \textcolor{green!50!black}{+4.96} \\

        Ministral3-14B
        & 50.48 & 53.45 & \textcolor{green!50!black}{+2.97}
        & 27.45 & 32.23 & \textcolor{green!50!black}{+4.78}
        & 49.71 & 53.22 & \textcolor{green!50!black}{+3.51} \\

        \bottomrule
    \end{tabular}
\end{table*}

\noindent\textbf{Overall Performance across Models and Benchmarks.} We train and evaluate \our{} separately on MMLU-Pro, SuperGPQA, and WildSci using seven LLMs. As shown in Tab. \ref{tab:main_results}, \our{} improves performance across all evaluated models and benchmarks, with gains spanning model scales from 3B to 32B parameters. The improvements extend beyond the Qwen family to the Ministral family, indicating that the effectiveness of  \our{} is not tied to a specific model family or task. This broad applicability is attributed to the query-adaptive design of \our{}, which generates neuron-wise modulation from query-induced internal activations without relying on fixed steering patterns or modifying the base model parameters. These results demonstrate the broad applicability of query-adaptive FFN neuron modulation across model scales, architectures, and task domains.

\begin{table*}[!t]
\centering
\scriptsize
\setlength{\tabcolsep}{2.6pt}
\renewcommand{\arraystretch}{1.15}

\caption{
In-domain performance and cross-benchmark generalization of Q-NeuS trained on MMLU-Pro.
Each cell reports accuracy (\%), with the change over the base model shown in parentheses.
}
\label{tab:generalization}

\resizebox{\textwidth}{!}{
\begin{tabular}{lcccccccc}
\toprule

\multirow{2}{*}{\textbf{Model}}
& \multicolumn{1}{c}{\textbf{In-Domain}}
& \multicolumn{7}{c}{\textbf{Cross-Benchmark Generalization}}
\\

\cmidrule(lr){2-2}
\cmidrule(lr){3-9}

& \textbf{MMLU-Pro}
& \textbf{GPQA-M}
& \textbf{GPQA-D}
& \textbf{SuperGPQA}
& \textbf{LiveBench}
& \textbf{AGIEval}
& \textbf{GSM8K}
& \textbf{ARC-C}
\\

\midrule

Qwen2.5-3B
& 38.37 {\scriptsize(\textcolor{green!50!black}{+6.94})}
& 33.04 {\scriptsize(\textcolor{green!50!black}{+2.91})}
& 36.87 {\scriptsize(\textcolor{green!50!black}{+4.55})}
& 19.39 {\scriptsize(\textcolor{green!50!black}{+0.13})}
& 24.15 {\scriptsize(\textcolor{red!70!black}{-0.48})}
& 52.32 {\scriptsize(\textcolor{green!50!black}{+0.31})}
& 81.20 {\scriptsize(\textcolor{red!70!black}{-2.27})}
& 82.76 {\scriptsize(\textcolor{green!50!black}{+0.17})}
\\

Qwen3-4B
& 45.93 {\scriptsize(\textcolor{green!50!black}{+1.07})}
& 36.61 {\scriptsize(0.00)}
& 39.90 {\scriptsize(\textcolor{green!50!black}{+2.53})}
& 27.72 {\scriptsize(\textcolor{green!50!black}{+1.30})}
& 55.06 {\scriptsize(\textcolor{red!70!black}{-0.26})}
& 58.29 {\scriptsize(\textcolor{red!70!black}{-0.63})}
& 90.75 {\scriptsize(\textcolor{green!50!black}{+0.30})}
& 89.59 {\scriptsize(\textcolor{red!70!black}{-0.60})}
\\

Qwen3-8B
& 48.53 {\scriptsize(\textcolor{green!50!black}{+4.33})}
& 36.38 {\scriptsize(\textcolor{green!50!black}{+0.22})}
& 36.87 {\scriptsize(\textcolor{green!50!black}{+3.03})}
& 29.54 {\scriptsize(\textcolor{green!50!black}{+1.11})}
& 51.96 {\scriptsize(\textcolor{green!50!black}{+0.63})}
& 59.28 {\scriptsize(\textcolor{red!70!black}{-0.37})}
& 93.25 {\scriptsize(\textcolor{green!50!black}{+0.60})}
& 91.64 {\scriptsize(\textcolor{green!50!black}{+0.68})}
\\

Qwen3-14B
& 54.03 {\scriptsize(\textcolor{green!50!black}{+3.14})}
& 40.62 {\scriptsize(\textcolor{green!50!black}{+2.45})}
& 42.93 {\scriptsize(\textcolor{green!50!black}{+4.04})}
& 32.71 {\scriptsize(\textcolor{green!50!black}{+3.23})}
& 57.57 {\scriptsize(\textcolor{green!50!black}{+0.36})}
& 65.55 {\scriptsize(\textcolor{green!50!black}{+0.14})}
& 94.54 {\scriptsize(\textcolor{green!50!black}{+0.23})}
& 92.66 {\scriptsize(\textcolor{green!50!black}{+0.42})}
\\

Qwen3-32B
& 59.11 {\scriptsize(\textcolor{green!50!black}{+4.55})}
& 42.41 {\scriptsize(\textcolor{red!70!black}{-1.12})}
& 43.43 {\scriptsize(0.00)}
& 33.46 {\scriptsize(\textcolor{green!50!black}{+4.14})}
& 58.08 {\scriptsize(\textcolor{red!70!black}{-0.61})}
& 69.41 {\scriptsize(\textcolor{green!50!black}{+0.04})}
& 95.45 {\scriptsize(\textcolor{green!50!black}{+0.53})}
& 94.80 {\scriptsize(\textcolor{green!50!black}{+0.26})}
\\

\midrule

Ministral3-8B
& 49.03 {\scriptsize(\textcolor{green!50!black}{+4.21})}
& 38.62 {\scriptsize(\textcolor{green!50!black}{+1.34})}
& 39.90 {\scriptsize(\textcolor{green!50!black}{+5.05})}
& 27.75 {\scriptsize(\textcolor{green!50!black}{+2.82})}
& 38.32 {\scriptsize(\textcolor{red!70!black}{-1.80})}
& 54.62 {\scriptsize(\textcolor{green!50!black}{+0.10})}
& 90.37 {\scriptsize(0.00)}
& 91.72 {\scriptsize(\textcolor{green!50!black}{+0.25})}
\\

Ministral3-14B
& 53.45 {\scriptsize(\textcolor{green!50!black}{+2.97})}
& 37.72 {\scriptsize(\textcolor{green!50!black}{+1.11})}
& 38.38 {\scriptsize(\textcolor{green!50!black}{+4.04})}
& 30.25 {\scriptsize(\textcolor{green!50!black}{+1.66})}
& 42.05 {\scriptsize(0.00)}
& 58.93 {\scriptsize(\textcolor{green!50!black}{+0.12})}
& 93.40 {\scriptsize(\textcolor{red!70!black}{-0.08})}
& 92.92 {\scriptsize(\textcolor{green!50!black}{+0.60})}
\\

\midrule

\textbf{Win / Tie / Loss}
& \textbf{7 / 0 / 0}
& \textbf{5 / 1 / 1}
& \textbf{6 / 1 / 0}
& \textbf{7 / 0 / 0}
& \textbf{2 / 1 / 4}
& \textbf{5 / 0 / 2}
& \textbf{4 / 1 / 2}
& \textbf{6 / 0 / 1}
\\

\bottomrule
\end{tabular}
}
\end{table*}

\begin{table*}[!t]
\centering
\small
\setlength{\tabcolsep}{4.5pt}
\renewcommand{\arraystretch}{1.15}

\caption{
Cross-benchmark transfer comparison between Q-NeuS and GRPO on Qwen3-8B.}
\label{tab:lora_comparison}

\resizebox{\textwidth}{!}{
\begin{tabular}{lcccccccc}
\toprule

\multirow{2}{*}{\textbf{Method}}
& \multicolumn{1}{c}{\textbf{In-Domain}}
& \multicolumn{7}{c}{\textbf{Cross-Benchmark Generalization}}
\\

\cmidrule(lr){2-2}
\cmidrule(lr){3-9}

& \textbf{MMLU-Pro}
& \textbf{GPQA-M}
& \textbf{GPQA-D}
& \textbf{SuperGPQA}
& \textbf{LiveBench}
& \textbf{AGIEval}
& \textbf{GSM8K}
& \textbf{ARC-C}
\\

\midrule

Base
& 44.20
& 36.16
& 33.84
& 28.43
& 51.33
& 59.65
& 92.65
& 90.96
\\

GRPO
& 48.33
& 35.71
& 33.84
& \textbf{30.77}
& 51.36
& \textbf{60.07}
& 93.18
& 91.21
\\

\textbf{Q-NeuS}
& \textbf{48.53}
& \textbf{36.38}
& \textbf{36.87}
& 29.54
& \textbf{51.96}
& 59.28
& \textbf{93.25}
& \textbf{91.64}
\\



\bottomrule
\end{tabular}
}
\end{table*}

\noindent\textbf{Cross-Benchmark Generalization.} We train \our{} only on MMLU-Pro and directly evaluate it on the other datasets in Tab. \ref{tab:generalization}. Across most benchmarks, performance improves or remains comparable to the base model, suggesting that the learned neuron modulation policy can transfer to other knowledge-intensive reasoning tasks. However, transfer gains on LiveBench are relatively limited, with only small improvements or occasional declines across models. This may be attributed to the larger distribution shift from MMLU-Pro: LiveBench includes tasks such as coding and data analysis, whose formats and required capabilities are less represented in the subject-oriented MMLU-Pro training data. As a result, both the candidate neuron space and the learned modulation policy may transfer less effectively to these different reasoning patterns. Meanwhile, Q-NeuS shows more consistent transfer on knowledge-intensive reasoning benchmarks such as GPQA and SuperGPQA, which are more closely aligned with the training distribution.


We further evaluate the transferability of \our{} and GRPO when both are trained only on the MMLU-Pro training set and directly applied to unseen benchmarks without target-domain training. As shown in Tab.~\ref{tab:lora_comparison}, both methods transfer beyond the source benchmark, while \our{} achieves better performance on most target benchmarks. We attribute this advantage to their adaptation mechanisms: GRPO learns fixed parameter updates from the source domain, while \our{} keeps the base LLM frozen and dynamically determines neuron-level interventions from the internal activations induced by each new query. It allows \our{} to adapt its intervention pattern to the knowledge and reasoning requirements of individual queries at inference time, rather than relying on a fixed adaptation learned from the source domain, thereby improving generalization to unseen task distributions.


\begin{table*}[!t]
\centering
\caption{
Subject-wise accuracy changes on MMLU-Pro with Qwen3-8B. The baseline row reports original accuracy, other rows show changes in percentage points (pp).
}
\label{tab:mmlupro_subject_delta}

\small
\setlength{\tabcolsep}{4.5pt}
\renewcommand{\arraystretch}{1.15}

\resizebox{\textwidth}{!}{
\begin{tabular}{lccccccccccccccc}
\toprule
\textbf{Method}
& \textbf{Bio.}
& \textbf{Bus.}
& \textbf{Chem.}
& \textbf{CS}
& \textbf{Econ.}
& \textbf{Eng.}
& \textbf{Health}
& \textbf{Hist.}
& \textbf{Law}
& \textbf{Math}
& \textbf{Other}
& \textbf{Phil.}
& \textbf{Phys.}
& \textbf{Psych.}
& \textbf{Avg.} 
\\
\midrule

Base
& 70.34
& 33.96
& 25.11
& 50.00
& 60.59
& 43.59
& 53.94
& 58.44
& 32.58
& 36.53
& 42.47
& 38.61
& 41.38
& 60.25
& 44.20
\\

Fix
& \textcolor{red!70!black}{-1.38}
& \textcolor{red!70!black}{-1.26}
& \textcolor{red!70!black}{-0.44}
& \textcolor{green!50!black}{+1.22}
& \textcolor{green!50!black}{+1.76}
& \textcolor{green!50!black}{+3.59}
& \textcolor{green!50!black}{+1.21}
& \textcolor{red!70!black}{-1.30}
& \textcolor{red!70!black}{-0.90}
& \textcolor{green!50!black}{+2.58}
& \textcolor{green!50!black}{+0.54}
& +0.00
& \textcolor{red!70!black}{-0.77}
& +0.00
& \textcolor{green!50!black}{+0.45}
\\

GRPO
& \textcolor{green!50!black}{+2.07}
& \textcolor{green!50!black}{+13.84}
& \textcolor{green!50!black}{+11.45}
& \textcolor{green!50!black}{+3.66}
& \textcolor{green!50!black}{+4.12}
& \textcolor{green!50!black}{+1.54}
& \textcolor{green!50!black}{+6.67}
& \textcolor{red!70!black}{-1.30}
& \textcolor{red!70!black}{-0.45}
& \textcolor{red!70!black}{-0.74}
& \textcolor{green!50!black}{+5.91}
& \textcolor{green!50!black}{+2.97}
& \textcolor{green!50!black}{+3.07}
& \textcolor{green!50!black}{+4.35}
& \textcolor{green!50!black}{+4.13}
\\

\textbf{Ours}
& \textcolor{green!50!black}{+3.45}
& \textcolor{green!50!black}{+5.03}
& \textcolor{green!50!black}{+7.49}
& \textcolor{green!50!black}{+1.22}
& \textcolor{green!50!black}{+5.88}
& \textcolor{green!50!black}{+3.59}
& \textcolor{green!50!black}{+13.94}
& \textcolor{green!50!black}{+1.30}
& \textcolor{green!50!black}{+4.07}
& \textcolor{red!70!black}{-0.74}
& \textcolor{green!50!black}{+4.30}
& \textcolor{green!50!black}{+2.97}
& \textcolor{green!50!black}{+2.30}
& \textcolor{green!50!black}{+5.59}
& \textbf{\textcolor{green!50!black}{+4.33}}
\\

\bottomrule
\end{tabular}
}
\end{table*}

\noindent\textbf{Comparison with Fixed Intervention and GRPO.} Tab. \ref{tab:mmlupro_subject_delta} presents the subject-wise comparison on MMLU-Pro. The fixed intervention strategy yields limited and unstable gains. In comparison, \our{} improves 13 of the 14 subjects and achieves an average gain of $+4.33$ percentage points. It indicates that predefined neuron selection, modulation strengths, and layer selection may not adequately accommodate the heterogeneous requirements of different subjects, whereas query-adaptive modulation provides broader improvements. Compared with GRPO, \our{} improves performance on a larger number of subjects. While GRPO learns a shared set of parameter updates across inputs, \our{} keeps the base LLM frozen and dynamically adjusts neuron-level modulation according to each query. These results demonstrate that query-adaptive neuron steering achieves competitive overall performance with more robust improvements across heterogeneous subjects.

\subsection{Ablation Studies}
\label{sec:ablation_studies}

We evaluate two key design choices in \our{}: the Neu-GRPO reward design and the neuron sampling ratio. All ablation experiments use Qwen3-8B on MMLU-Pro.

\begin{table}[!t]
\centering
\caption{
Neu-GRPO reward ablation on MMLU-Pro with Qwen3-8B. Baseline: 44.20\%.
}
\label{tab:reward_ablation}

\small
\setlength{\tabcolsep}{7pt}
\renewcommand{\arraystretch}{1.12}

\begin{tabular}{lccccc}
\toprule
\textbf{Variant}
& $\lambda_{\mathrm{c}}$
& $\lambda_{\mathrm{m}}$
& $\lambda_{\mathrm{s}}$
& \textbf{Acc. (\%)}
& $\boldsymbol{\Delta}$ \textbf{(pp)} \\
\midrule

w/o $R_{\mathrm{margin}}$
& 1.0 & 0 & 0.1
& 48.04
& \textcolor{green!50!black}{+3.84} \\

w/o $R_{\mathrm{strength}}$
& 1.0 & 1.0 & 0
& 47.83
& \textcolor{green!50!black}{+3.63} \\

w/o $R_{\mathrm{margin}}, R_{\mathrm{strength}}$
& 1.0 & 0 & 0
& 47.50
& \textcolor{green!50!black}{+3.30} \\

\midrule

\textbf{Full Neu-GRPO}
& \textbf{1.0} & \textbf{1.0} & \textbf{0.1}
& \textbf{48.53}
& \textbf{\textcolor{green!50!black}{+4.33}} \\

\bottomrule
\end{tabular}
\end{table}

\noindent\textbf{Effect of Reward Design.} The Neu-GRPO reward consists of correctness, margin, and intervention-strength terms. To evaluate their contributions, we remove $R_{\mathrm{margin}}$, $R_{\mathrm{strength}}$, or both and compare the resulting variants with the full objective. As shown in Tab.~\ref{tab:reward_ablation}, the correctness reward alone improves performance, while the full reward achieves the best result. This indicates that the auxiliary terms provide complementary optimization signals beyond correctness. Specifically, the margin reward distinguishes rollouts with identical correctness outcomes through their prediction margins, while the strength penalty discourages excessive neuron modulation. Together, they provide a more informative and stable learning signal for optimizing the neuron intervention policy.

\begin{table}[!t]
\centering
\caption{
Effect of the neuron sampling ratio on MMLU-Pro with Qwen3-8B.
}
\label{tab:sampling_ratio}

\small
\setlength{\tabcolsep}{4.5pt}
\renewcommand{\arraystretch}{1.15}

\resizebox{\columnwidth}{!}{
\begin{tabular}{lcccccccccc}
\toprule
\textbf{Sampling Ratio}
& Baseline (0\%)
& 10\%
& 20\%
& 30\%
& 40\%
& 50\%
& \textbf{60\%}
& 70\%
& 80\%
& 90\%
\\
\midrule

\textbf{Acc. (\%)}
& 44.20
& 46.76
& 47.13
& 47.38
& 48.24
& 48.16
& \textbf{48.53}
& 48.29
& 48.29
& 46.80
\\

$\boldsymbol{\Delta}$ \textbf{(pp)}
& +0.00
& \textcolor{green!50!black}{+2.56}
& \textcolor{green!50!black}{+2.93}
& \textcolor{green!50!black}{+3.18}
& \textcolor{green!50!black}{+4.04}
& \textcolor{green!50!black}{+3.96}
& \textbf{\textcolor{green!50!black}{+4.33}}
& \textcolor{green!50!black}{+4.09}
& \textcolor{green!50!black}{+4.09}
& \textcolor{green!50!black}{+2.60}
\\

\bottomrule
\end{tabular}
}
\end{table}

\noindent\textbf{Effect of Neuron Sampling Ratio.} During Neu-GRPO training, we construct intervention rollouts by sampling different neuron subsets from the candidate neuron space. To study the effect of the sampling ratio, we vary it from 10\% to 90\%. As shown in Tab. \ref{tab:sampling_ratio}, Q-NeuS outperforms the base model across all sampling ratios, demonstrating robustness to this hyperparameter. The accuracy improves as the sampling ratio increases and remains relatively stable between 40\% and 80\%, indicating that the benefit of increasing neuron coverage gradually saturates. The best result is achieved at 60\%. However, increasing the ratio to 90\% leads to a clear performance drop, suggesting that excessive neuron sampling may reduce the diversity of intervention configurations across rollouts.


\begin{figure*}[!t]
    \centering
    \includegraphics[width=\textwidth]{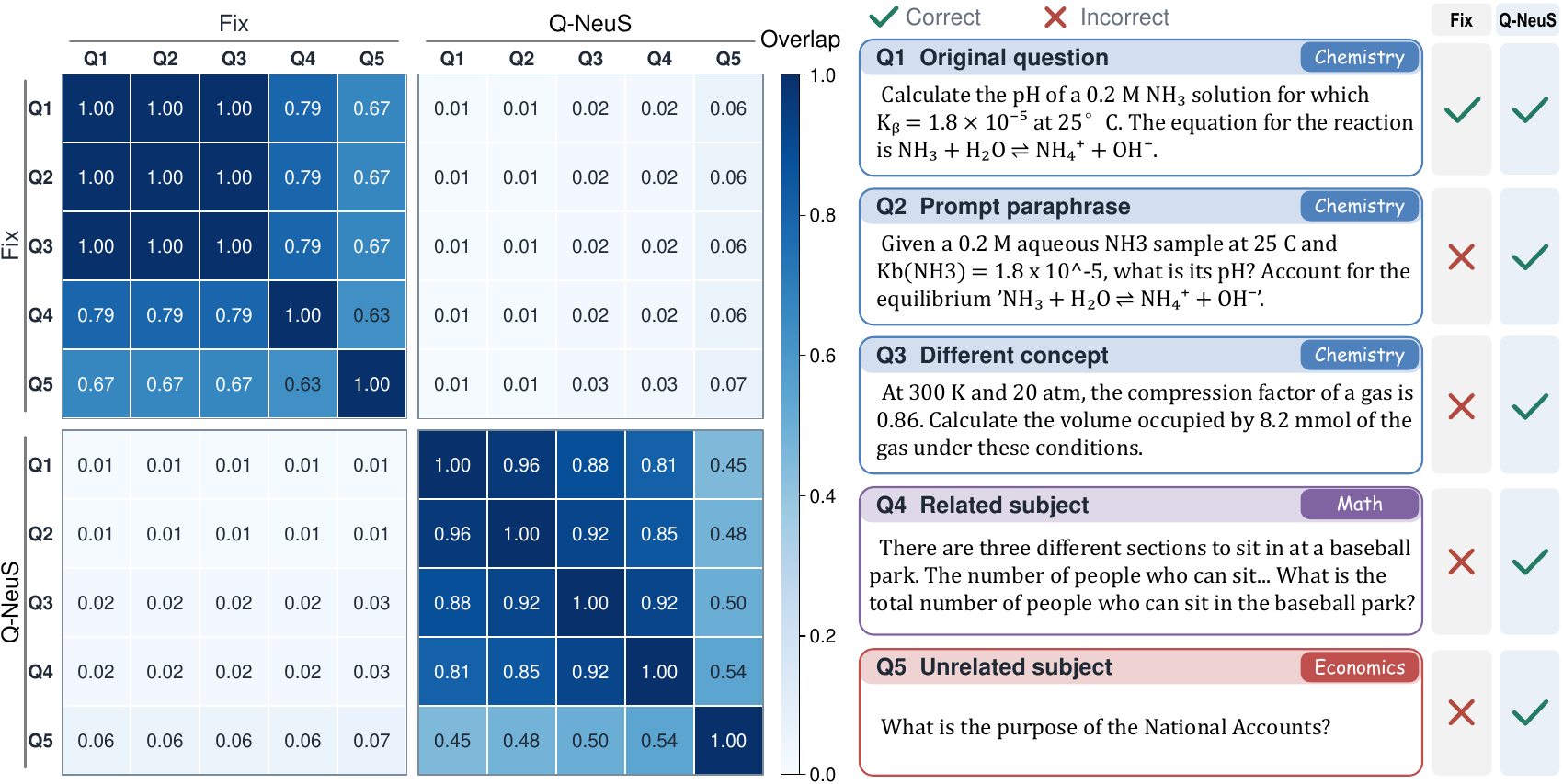}
    \caption{Case study of Fix and \our{} across queries with different semantic relationships.}
    \label{case_study}
\end{figure*}

\subsection{Case Study}

We compare Fix and \our{} on five queries: an original chemistry question (Q1), its paraphrase (Q2), a chemistry question involving a different concept (Q3), a question from a related subject, mathematics (Q4), and a question from an unrelated subject, economics (Q5). 
Fig.~\ref{case_study} presents the overlap among the top-100 most strongly modulated neurons together with the corresponding answer correctness. Fix applies a fixed subject-specific intervention, resulting in identical top-100 modulated neurons for Q1--Q3. Although related subjects exhibit higher overlap than unrelated ones, the differences remain limited. It indicates that Fix captures coarse subject-level relationships but cannot adapt its intervention to individual queries within the same subject.

\our{} exhibits three notable properties: 1) It produces query-specific neuron patterns even within the same subject: the overlap between Q1 and Q3 is 0.88 despite both being chemistry questions. 2) It preserves a high overlap of 0.96 between Q1 and its paraphrase Q2 and answers both correctly, whereas Fix fails on Q2 despite succeeding on Q1, demonstrating robustness to prompt reformulation. 
3) The selected neuron sets reflect semantic relationships across queries: relative to Q1, the overlap is 0.81 for the related-subject query Q4 but drops to 0.45 for the unrelated-subject query Q5, yielding a clearer distinction than Fix. \our{} answers all five queries correctly, while Fix succeeds only on Q1. These results illustrate that \our{} can adjust neuron-level interventions to individual queries while preserving similar modulation patterns for semantically equivalent inputs. Appendix \ref{app:case_study} further shows that these query-specific overlap patterns vary across Transformer layers and exhibit clear layer-dependent structure, consistent with our preliminary findings.
\section{Conclusion}

In this work, we investigate query-adaptive neuron steering for LLMs. Our preliminary study shows that highly activated neurons exhibit subject- and layer-dependent patterns, and that amplifying subject-relevant neurons can improve model performance. Motivated by these findings, we propose \our{}, which dynamically determines which neurons to modulate and by how much from query-induced internal activations. \our{} combines a subject-guided candidate neuron space with a lightweight Query-Adaptive Gate and introduces Neu-GRPO to optimize neuron modulation through group-relative feedback while keeping the base LLM frozen. Extensive experiments across seven LLMs and multiple benchmarks demonstrate that \our{} improves performance across model scales and architectures and transfers effectively to multiple unseen benchmarks with diverse knowledge and reasoning requirements. These results show that query-induced internal activations can provide effective signals for adaptive intervention, enabling LLMs to better elicit their existing capabilities without updating the base model parameters or requiring task labels at inference time.
\bibliographystyle{unsrtnat}

\bibliography{ref}

\clearpage
\newpage
\beginappendix

\section{Additional Analysis of the Preliminary Study}
\label{app:preliminary}

\subsection{Neuron Overlap Patterns across Layers}
\label{app:overlap_across_layers}

\begin{figure*}[t]
    \centering
    \captionsetup[subfigure]{skip=2pt}

    \begin{subfigure}[t]{0.49\textwidth}
        \centering
        \includegraphics[width=\linewidth]{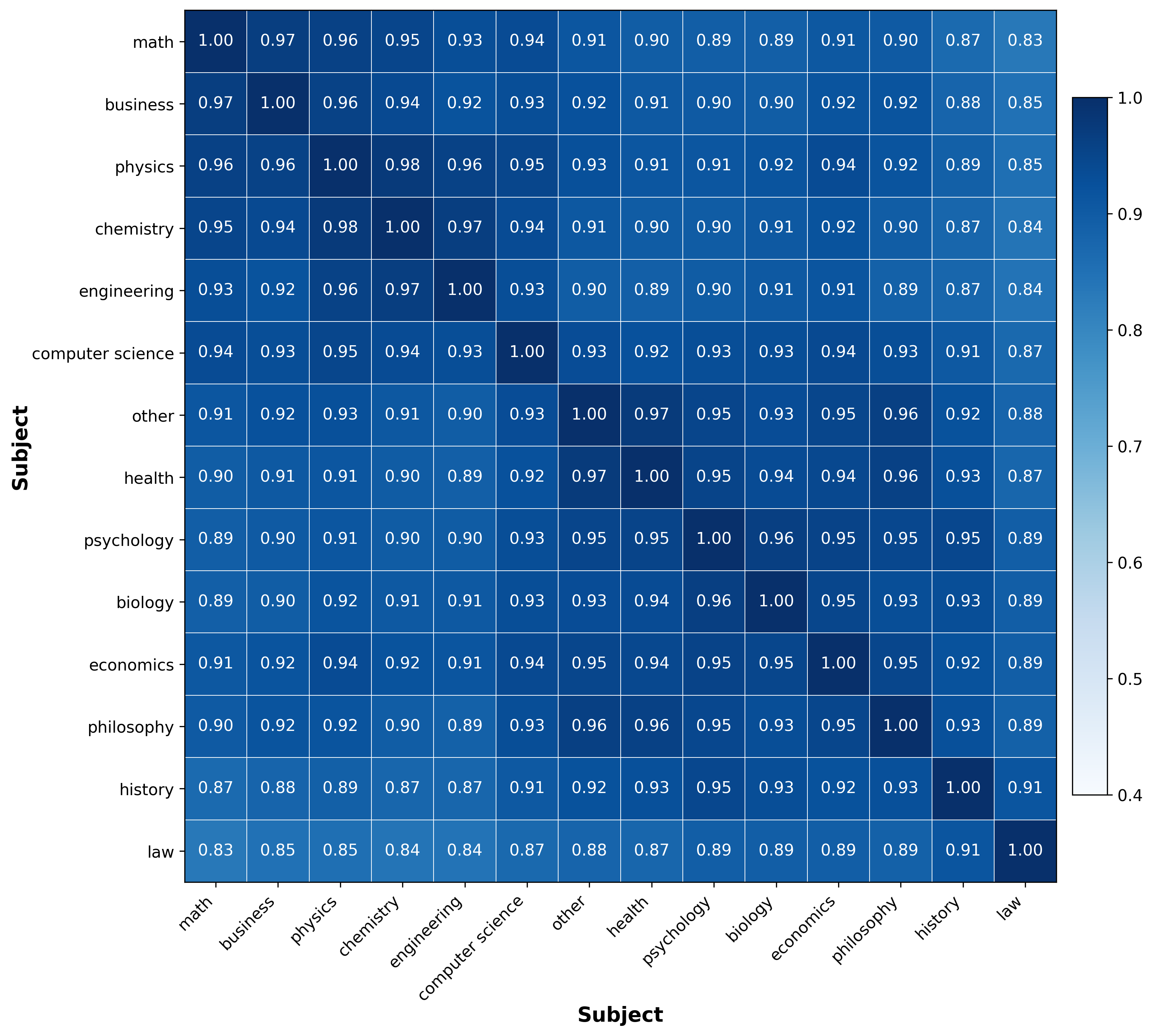}
        \caption{Layer 5}
        \label{fig:overlap_layer5}
    \end{subfigure}%
    \hspace{0.005\textwidth}%
    \begin{subfigure}[t]{0.49\textwidth}
        \centering
        \includegraphics[width=\linewidth]{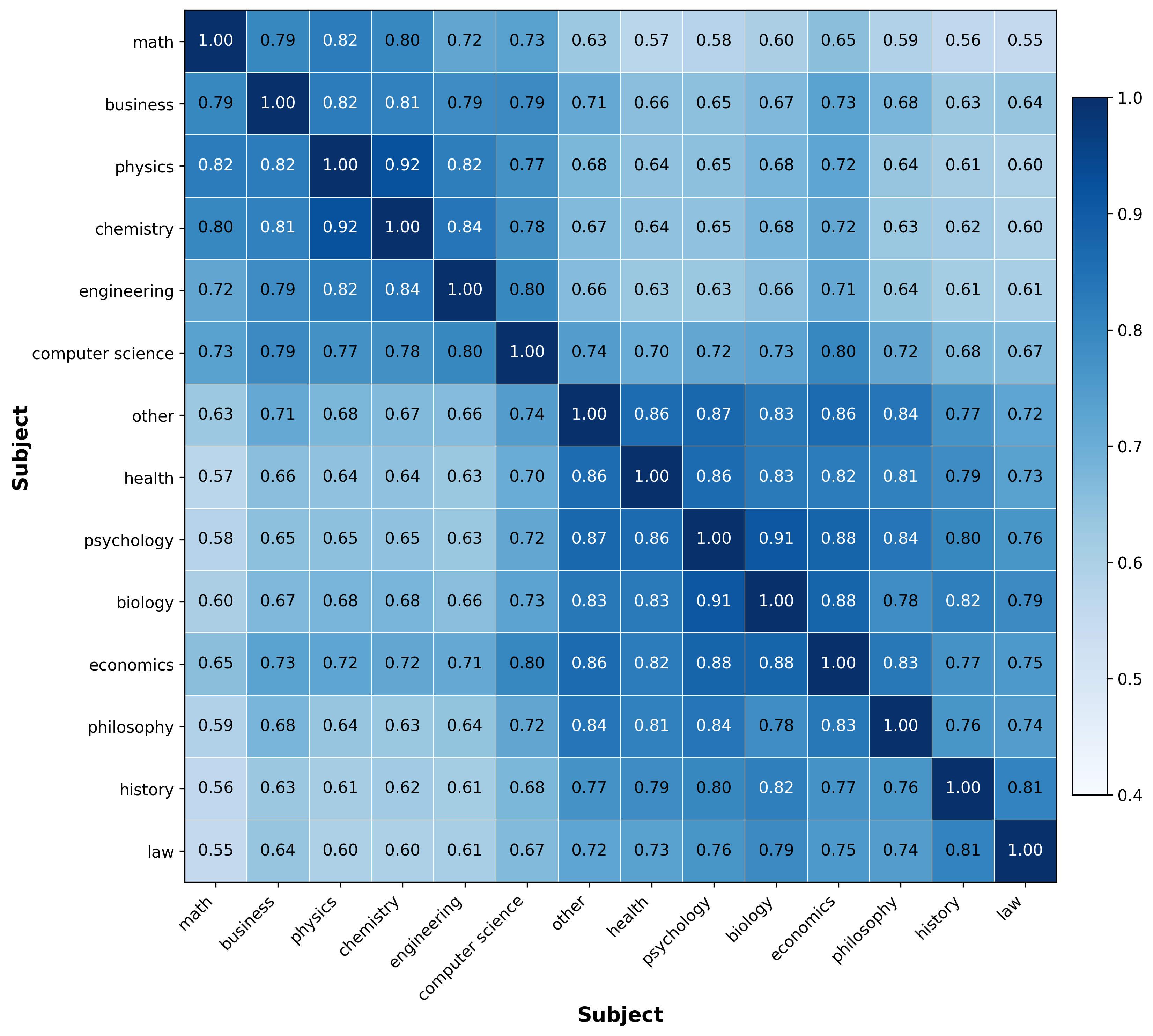}
        \caption{Layer 15}
        \label{fig:overlap_layer15}
    \end{subfigure}

    \vspace{0.4em}

    \begin{subfigure}[t]{0.49\textwidth}
        \centering
        \includegraphics[width=\linewidth]{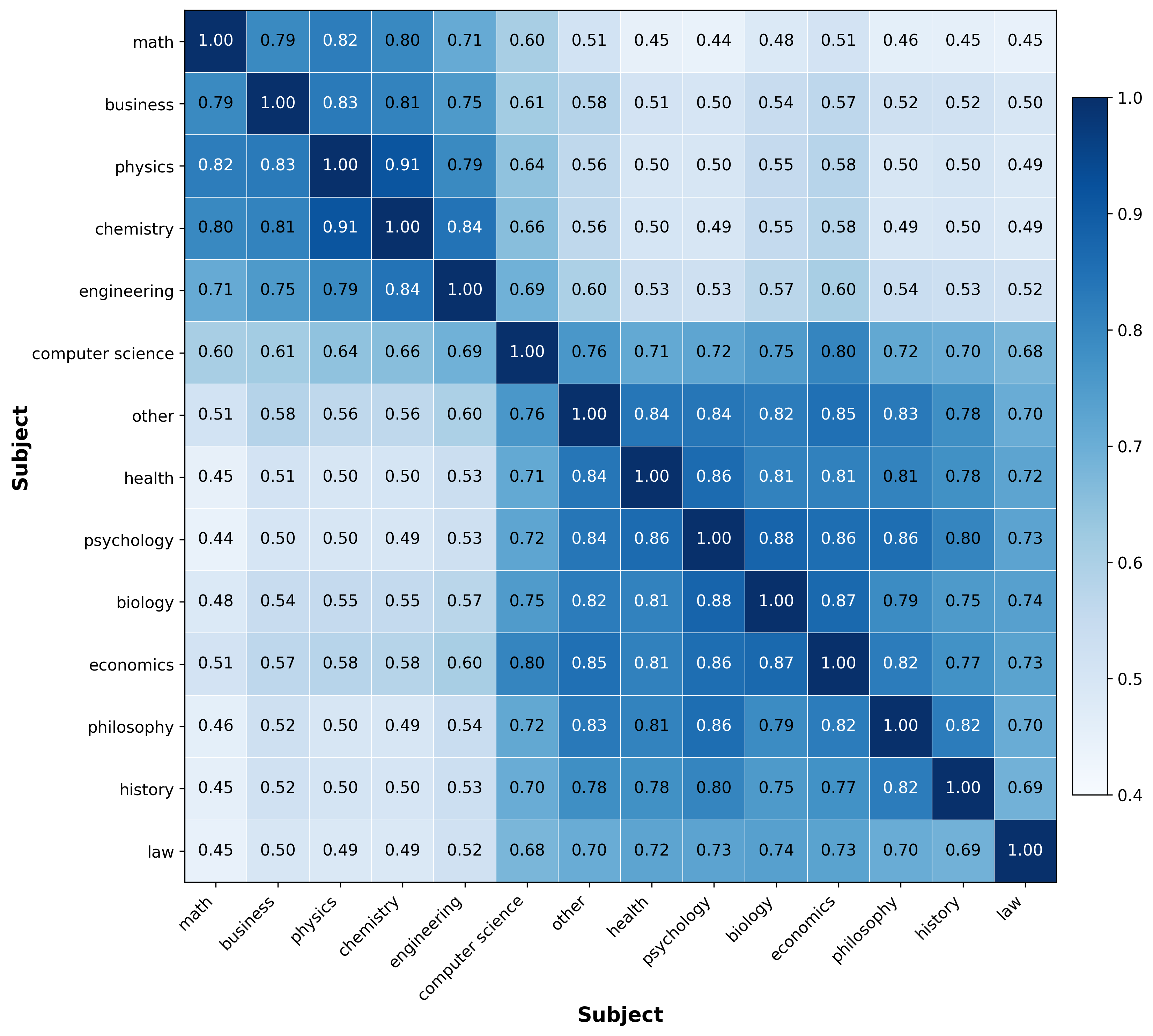}
        \caption{Layer 25}
        \label{fig:overlap_layer25}
    \end{subfigure}%
    \hspace{0.005\textwidth}%
    \begin{subfigure}[t]{0.49\textwidth}
        \centering
        \includegraphics[width=\linewidth]{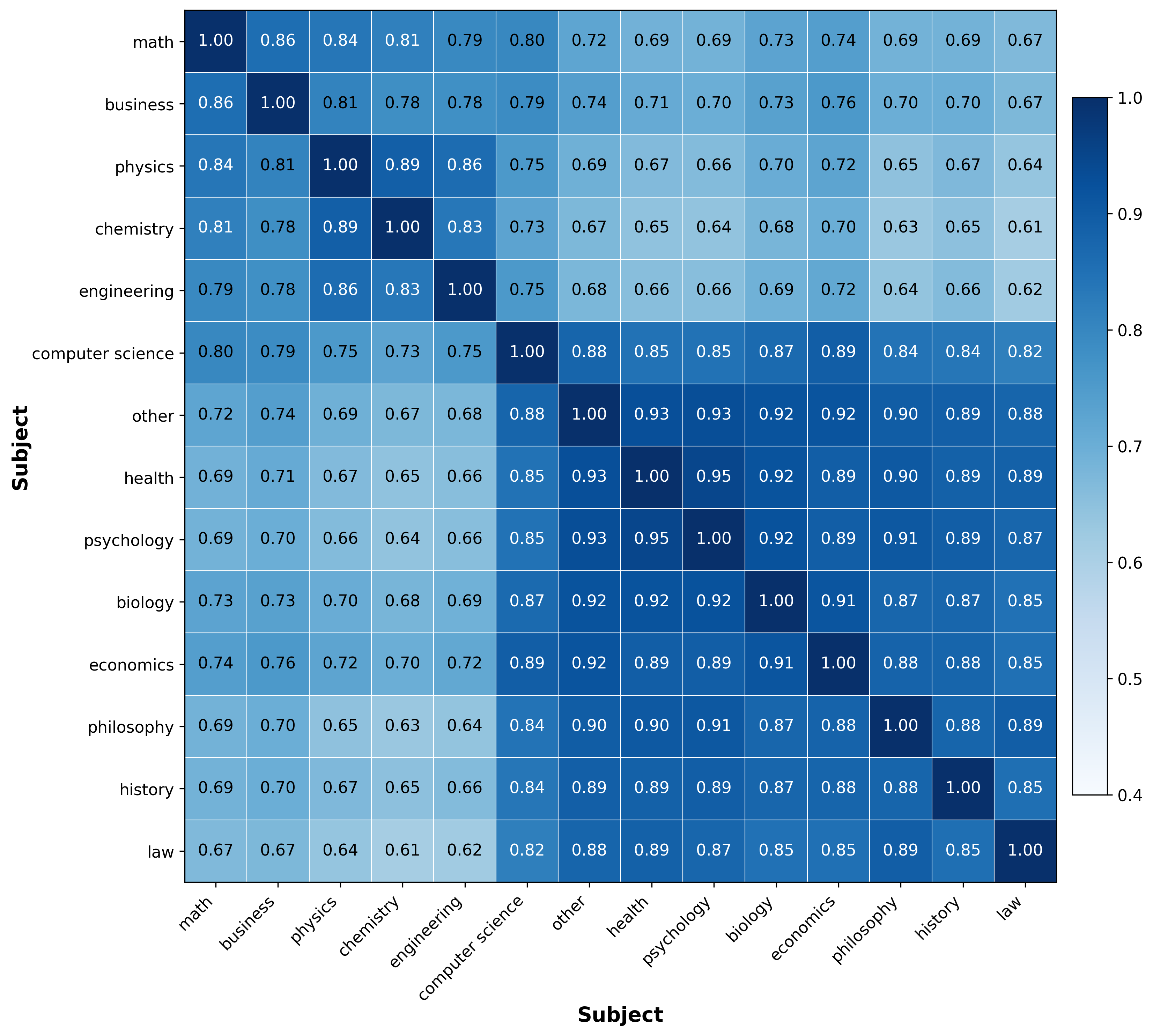}
        \caption{Layer 35}
        \label{fig:overlap_layer35}
    \end{subfigure}

    \caption{
    Neuron overlap matrices across representative Transformer layers.
    Each matrix reports the pairwise overlap rates of the Top-500 highly activated FFN neurons across MMLU-Pro subjects.
    }
    \label{fig:overlap_layers}
\end{figure*}

To further analyze the organization of subject-related neurons across different layers, we compute the overlap matrices of the Top-500 highly activated neurons across subjects at several representative layers, as shown in Fig.~\ref{fig:overlap_layers}. Overall, different layers exhibit clearly distinct organizational patterns.

In shallower layers, such as layer 5, nearly all subject pairs exhibit relatively high overlap rates, suggesting that highly activated neurons in shallow layers largely reflect shared representations across subjects, with limited subject-specific differentiation. In intermediate layers, such as layers 15 and 25, subjects with similar semantic content or reasoning patterns tend to exhibit higher neuron overlap, whereas more dissimilar subjects show substantially lower overlap, forming clearer block-like structures. This differentiation is most pronounced at layer 25, indicating stronger subject-related neuron organization in intermediate layers. In deeper layers, such as layer 35, the overall overlap across subjects increases again, while the block structure becomes less distinct.

These observations further indicate that highly activated neurons exhibit different organizational structures across layers.

\subsection{Overall Activation Magnitude across Layers}
\label{app:activation_magnitude}

\begin{figure*}[t]
    \centering
    \includegraphics[width=\textwidth]{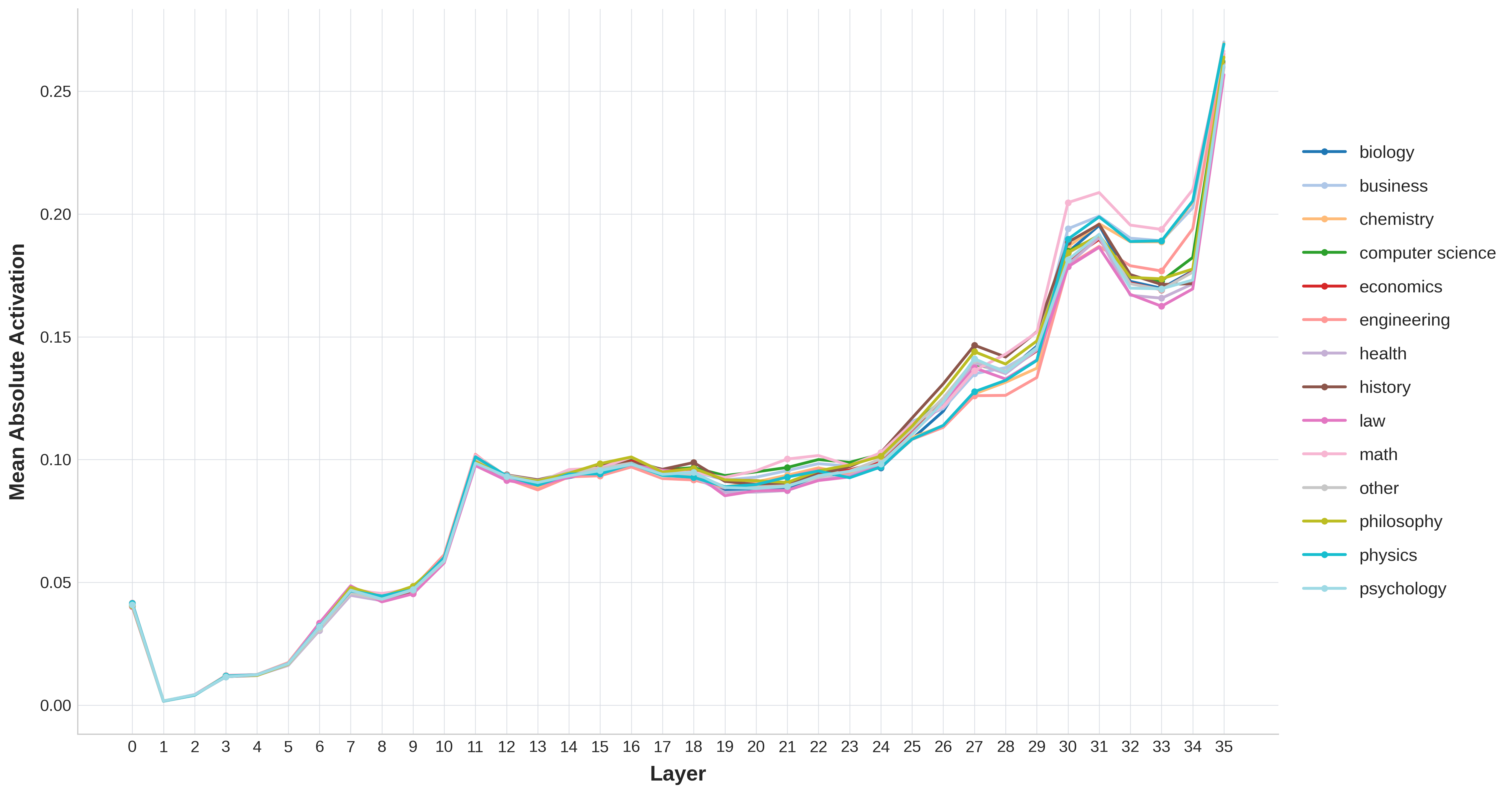}
    \caption{
    Mean absolute FFN neuron activations across Transformer layers for different MMLU-Pro subjects.
    }
    \label{fig:activation_magnitude}
\end{figure*}

To further examine how internal activations vary with network depth, we compute the mean absolute activation of all FFN neurons at each Transformer layer for samples from different subjects. As shown in Fig.~\ref{fig:activation_magnitude}, different subjects exhibit highly similar layer-wise trends: the overall activation magnitude is relatively low in shallow layers, increases gradually through the intermediate layers, and becomes stronger in deeper layers.

In contrast, the differences in mean activation magnitude across subjects within the same layer are relatively small, with noticeable divergence appearing only in some deeper layers. This suggests that subject-related differences are not primarily reflected in the overall activation magnitude. Combined with the neuron overlap analysis in the previous subsection, these results suggest that subject-related differences are more likely to manifest in the specific composition and distribution patterns of activated neurons.

\subsection{Layer-wise Intervention Effects}
\label{app:layer_intervention}

We further analyze the intervention effects across different layers in the preliminary study. During validation, the 36 Transformer layers exhibit substantially different responses to neuron intervention. Among them, 21 layers achieve positive accuracy gains, while the remaining layers show only limited improvements or performance degradation. Among all single-layer interventions, layer 18 yields the largest improvement, increasing accuracy by 1.17 percentage points.

In addition to the layer-selection results reported in the main text, we further examine the setting without layer selection. Specifically, we remove the validation-based layer selection and directly apply joint neuron intervention to all 36 Transformer layers. Under this setting, the model accuracy decreases from the baseline of 31.56\% to 29.41\%, corresponding to a drop of 2.15 percentage points. In contrast, jointly intervening on the 21 layers selected on the validation set achieves an accuracy of 32.92\%.

This comparison shows that even though intervention on some individual layers can yield positive gains, indiscriminately extending the intervention to all layers can substantially degrade overall performance. Therefore, simply increasing the intervention coverage does not necessarily lead to better performance.

\section{Parameter Efficiency of the Query-Adaptive Gate}
\label{app:parameter_efficiency}

Q-NeuS employs a single Query-Adaptive Gate shared across all Transformer layers, while using low-dimensional layer embeddings to preserve layer-specific information. As shown in Tab.~\ref{tab:gate_parameters}, across the seven models evaluated in this work, the Query-Adaptive Gate contains only 39.92M--104.95M parameters, corresponding to 79.85--209.91MB of BF16 weights. This accounts for only 0.32\%--1.46\% of the parameters of the corresponding text backbone models. As the backbone model becomes larger, the relative parameter overhead generally decreases. For example, the Query-Adaptive Gate accounts for 0.99\%, 0.62\%, 0.48\%, and 0.32\% of the parameters of Qwen3-4B, Qwen3-8B, Qwen3-14B, and Qwen3-32B, respectively.

Cross-layer sharing is a key factor in reducing the additional parameter cost. If an independent Query-Adaptive Gate were introduced for each Transformer layer, the number of gating parameters would increase by approximately 34--64$\times$ across the evaluated models. Therefore, the current design substantially reduces the parameter and storage overhead of the additional module while retaining layer-specific conditioning.

\begin{table}[t]
\centering
\small
\setlength{\tabcolsep}{6pt}
\renewcommand{\arraystretch}{1.1}
\caption{
Parameter overhead of the shared Query-Adaptive Gate across different backbone models.
Weight storage is reported in BF16.
}
\label{tab:gate_parameters}
\begin{tabular}{lccc}
\toprule
\textbf{Model}
& \textbf{Gate Params}
& \textbf{Gate / Backbone}
& \textbf{BF16 Weight Size}
\\
\midrule

Qwen2.5-3B
& 45.17M
& 1.46\%
& 90.34MB
\\

Qwen3-4B
& 39.92M
& 0.99\%
& 79.85MB
\\

Qwen3-8B
& 50.41M
& 0.62\%
& 100.83MB
\\

Qwen3-14B
& 71.39M
& 0.48\%
& 142.78MB
\\

Qwen3-32B
& 104.95M
& 0.32\%
& 209.91MB
\\

Ministral3-8B
& 58.80M
& 0.69\%
& 117.61MB
\\

Ministral3-14B
& 67.19M
& 0.50\%
& 134.39MB
\\

\bottomrule
\end{tabular}
\end{table}

\section{Additional Case Study Analysis}
\label{app:case_study}

\begin{figure*}[t]
    \centering
    \captionsetup[subfigure]{skip=2pt}

    \begin{subfigure}[t]{0.325\textwidth}
        \centering
        \includegraphics[height=0.19\textheight]{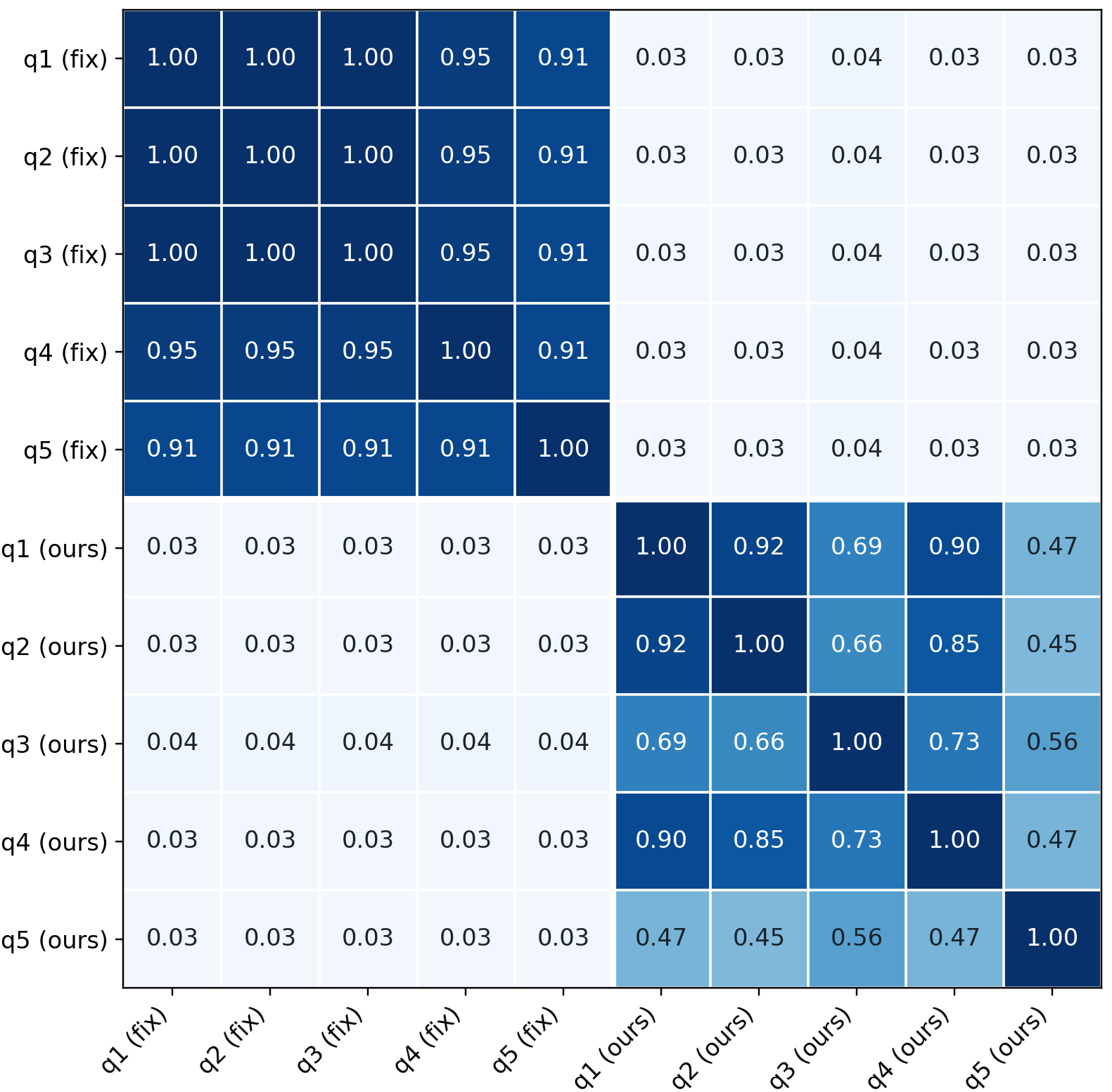}
        \caption{Layer 1}
        \label{fig:case_layer01}
    \end{subfigure}%
    \hspace{0.005\textwidth}%
    \begin{subfigure}[t]{0.325\textwidth}
        \centering
        \includegraphics[height=0.19\textheight]{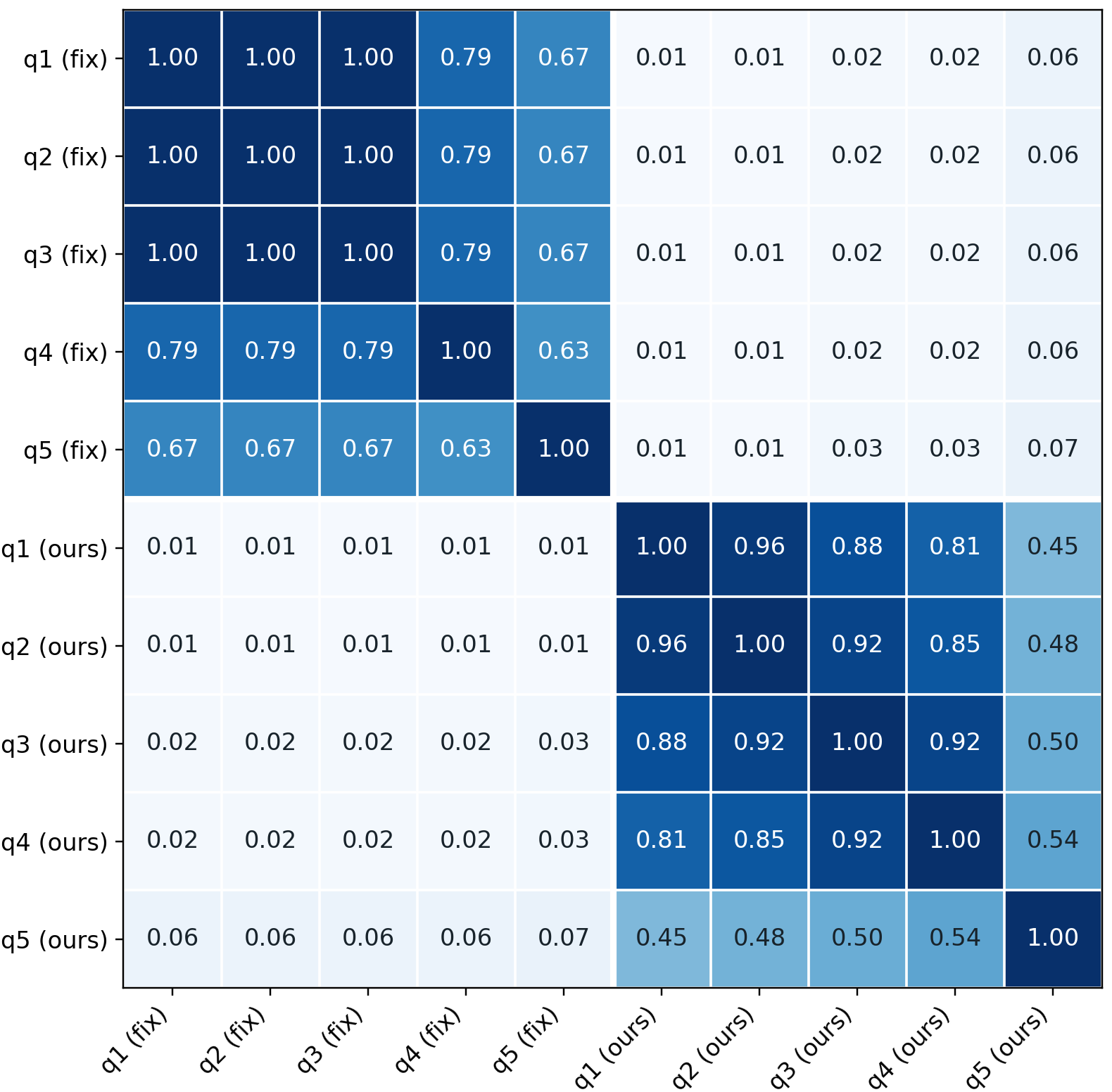}
        \caption{Layer 15}
        \label{fig:case_layer15}
    \end{subfigure}%
    \hspace{0.005\textwidth}%
    \begin{subfigure}[t]{0.325\textwidth}
        \centering
        \includegraphics[height=0.19\textheight]{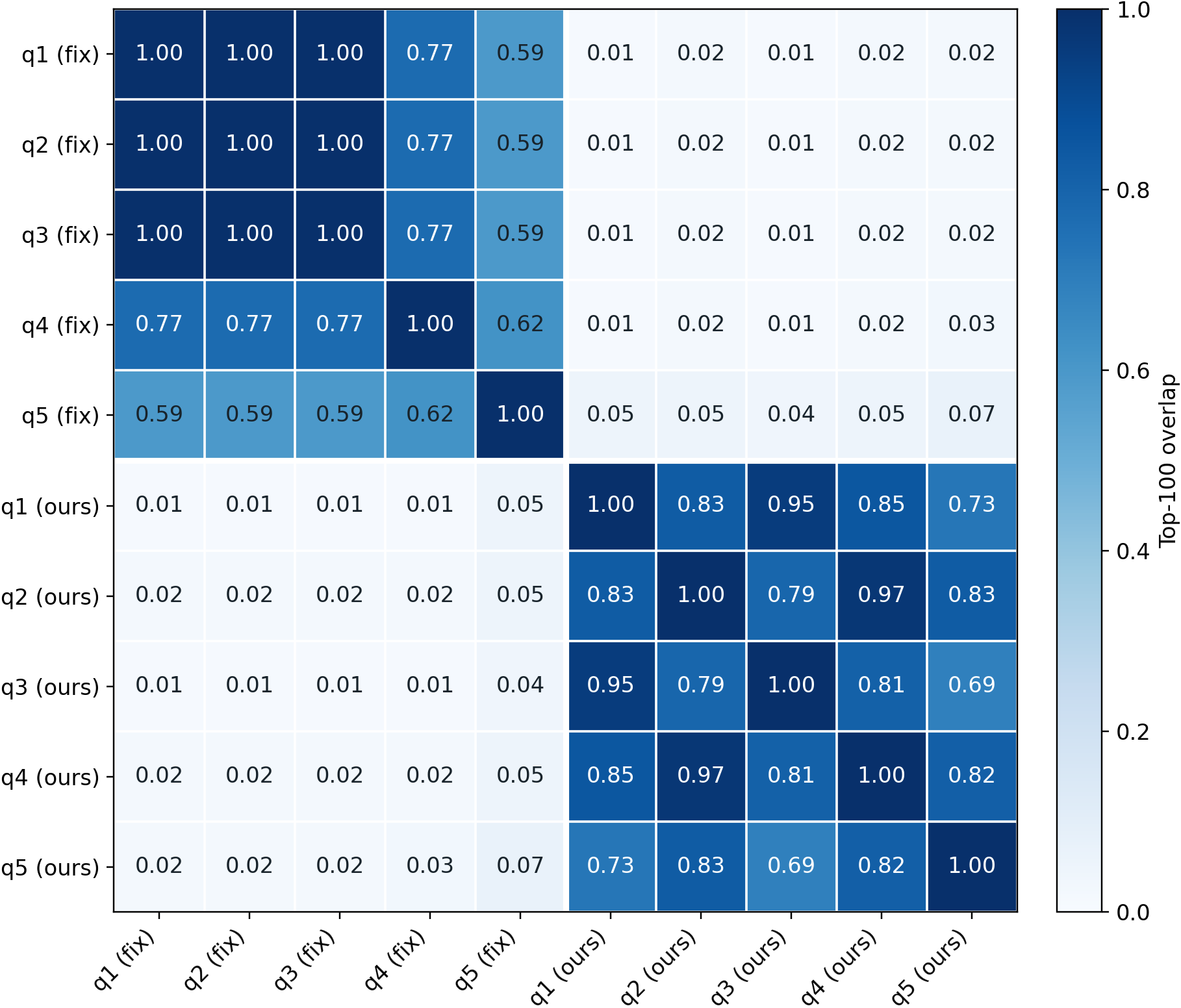}
        \caption{Layer 18}
        \label{fig:case_layer18}
    \end{subfigure}

    \caption{
    Top-100 neuron overlap matrices of Fix and Q-NeuS across different Transformer layers for the case-study queries.
    }
    \label{fig:case_layers}
\end{figure*}

We further analyze how the neuron overlap patterns in the case study vary across different Transformer layers. As shown in Fig.~\ref{fig:case_layers}, different layers exhibit distinct query-level overlap structures, indicating that the query-adaptive neuron patterns are clearly layer-dependent, consistent with the layer-dependent neuron organization observed in the preliminary study.

\end{document}